\documentclass[conference]{IEEEtran}
\IEEEoverridecommandlockouts
\usepackage{amsmath,amssymb,amsfonts, balance}
\usepackage{algorithmic}
\usepackage{graphicx}
\usepackage{textcomp}
\usepackage{hyperref}
\usepackage[usenames,dvipsnames]{xcolor}
\usepackage[numbers]{natbib}

\begin{document}

\title{GloSoFarID: Global multispectral dataset for Solar Farm IDentification in satellite imagery\\

}

\author{\IEEEauthorblockN{Zhiyuan Yang and Ryan Rad}
\IEEEauthorblockA{\textit{Khoury College of Computer Science, Northeastern University}\\
Vancouver, Canada \\
\{yang.zhiyu, r.rad\}@northeastern.edu}
}

\maketitle

\begin{abstract}
Solar Photovoltaic (PV) technology is increasingly recognized as a pivotal solution in the global pursuit of clean and renewable energy. This technology addresses the urgent need for sustainable energy alternatives by converting solar power into electricity without greenhouse gas emissions. It not only curtails global carbon emissions but also reduces reliance on finite, non-renewable energy sources. In this context, monitoring solar panel farms becomes essential for understanding and facilitating the worldwide shift toward clean energy. This study contributes to this effort by developing the first comprehensive global dataset of multispectral satellite imagery of solar panel farms. This dataset is intended to form the basis for training robust machine learning models, which can accurately map and analyze the expansion and distribution of solar panel farms globally. The insights gained from this endeavor will be instrumental in guiding informed decision-making for a sustainable energy future. \href{https://github.com/yzyly1992/GloSoFarID}{https://github.com/yzyly1992/GloSoFarID}
\end{abstract}

\begin{IEEEkeywords}
photovoltaic, solar panel farm, multispectral images, dataset, sustainability
\end{IEEEkeywords}

\section{Introduction}
Solar Photovoltaic (PV) technology has emerged as a key solution for meeting the global demand for clean and renewable energy. By harnessing solar power for electricity generation, solar PV plays a significant role in mitigating climate change and reducing pollution. Its ability to replace traditional power generation methods reduces global carbon emissions and lessens dependence on finite resources, paving the way for a sustainable future \cite{sustainability}.

A critical aspect of the global transition to clean energy is the precise monitoring of solar panel farms, which are integral to renewable energy infrastructure worldwide. Understanding the growth and distribution of these farms is essential for assessing progress towards a sustainable energy landscape \cite{trend}. Therefore, monitoring these farms is a pivotal step in comprehensively understanding the global adoption of clean energy resources.

In line with this goal, our project aims to make a significant contribution by creating a comprehensive dataset of multispectral satellite images of solar panel farms. This dataset will be a foundational resource for developing effective machine learning models for monitoring solar panel farms globally. The focus on using these models highlights the importance of accurate and advanced monitoring techniques in promoting a sustainable and cleaner energy future worldwide. This dataset is a pioneering effort, representing the first global compilation of solar panel farm data, and includes comprehensive multispectral satellite information. Our dataset enhances existing resources in three ways:
\begin{itemize}
    \item It employs mid-resolution data and includes rural areas globally, addressing previous limitations in geographic coverage.
    \item It provides multispectral satellite images with $13$ multispectral bands, offering more comprehensive information that comes in an easy-to-use format with a complementary how-to-use guide and script. 
    \item The dataset is updated with recent data from $2021$ to $2023$, doubling the size of the existing data sources and ensuring current and relevant insights into the global solar energy landscape.
\end{itemize}

\begin{figure*}[htbp]
\centerline{\includegraphics[width=\textwidth]{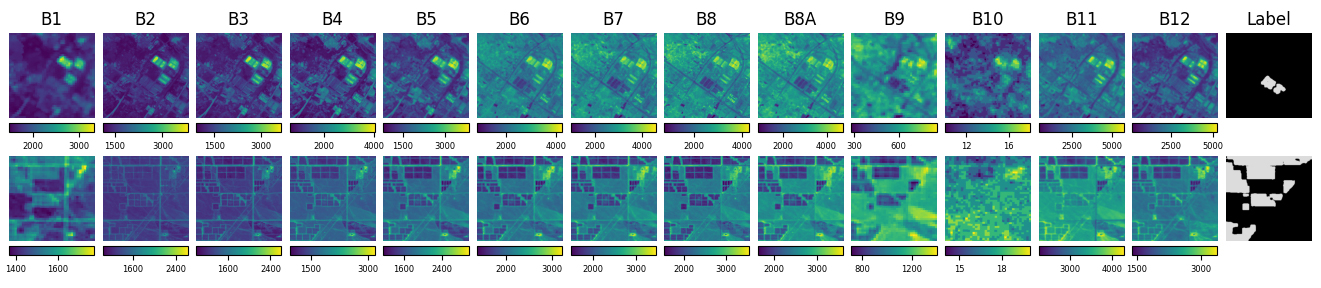}}
\caption{Examples from our GloSoFarID dataset. Each row represents one sample with all $13$ bands along with the ground truth mask where the gray color indicates the solar farm area. See TABLE \ref{tab:bands-info} for more details.}
\label{fig-data-sample}
\end{figure*}

\section{Dataset Construction}
\begin{figure}[htbp]
\centerline{\includegraphics[width=0.4\textwidth]{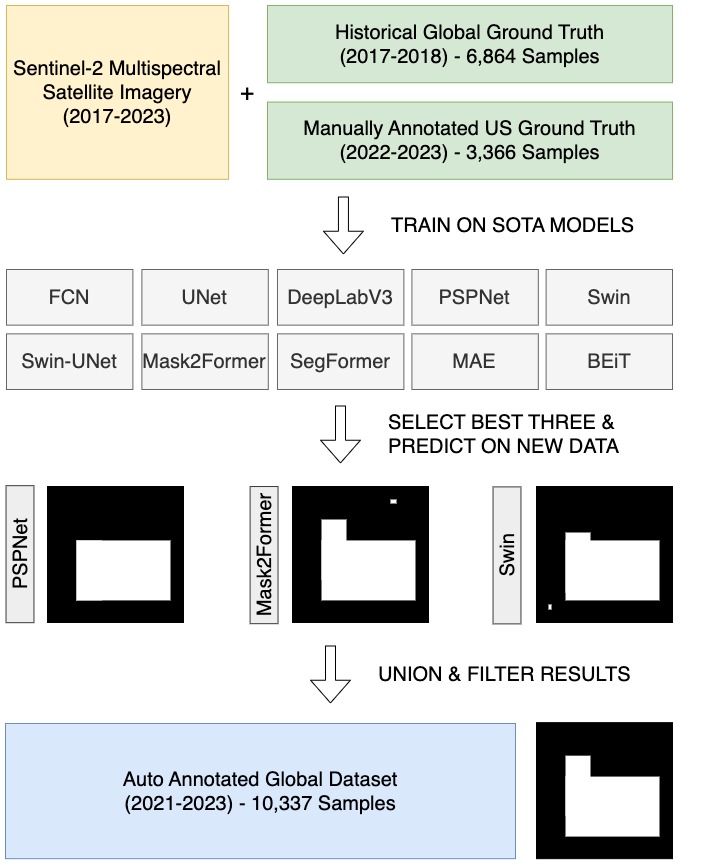}}
\caption{Dataset Construction Pipeline}
\label{fig-data-gen}
\end{figure}

To construct this dataset, we devised a methodical three-step process that incorporates modern techniques. As illustrated in Fig. \ref{fig-data-gen}, our approach begins with the creation of a robust initial training set, proceeds with the training of state-of-the-art (SOTA) models, and concludes by processing the combined outputs of the top-performing models to generate a novel dataset.

\subsection{Gather Initial Training Dataset}
Our first step involved aggregating data from diverse sources to assemble an initial dataset. This dataset comprises three primary segments: Historical Global Ground Truth from $2017$-$2018$ \cite{globalinventory}, Manually Annotated US Ground Truth from $2022$-$2023$ by our team using Google Earth Engine (GEE), and Sentinel-2 Multispectral Satellite Imagery spanning $2017$-$2023$ \cite{sentinel}. The integration of these datasets yielded a comprehensive initial training set of $10,230$ multispectral images, each with a resolution of $256 \times 256$ pixels and encompassing $13$ spectral bands. This rich and varied dataset lays the groundwork for our subsequent analytical endeavors.

\subsection{Train SOTA Models}
With the initial training set in place, we engaged several SOTA models, renowned for their performance, for training. Utilizing a batch size of $16$ over $50$ epochs, our models demonstrated exceptional accuracy in identifying solar panel farms from multispectral images. The top-performing model achieved an outstanding $96.47\%$ Intersection over Union (IoU) and a $98.2\%$ F-score, demonstrating its high accuracy in precisely segmenting solar panel areas within the imagery.

\subsection{Generate New Dataset}
Following the identification of the top three models based on predictive accuracy, we employed them to predict and label data for the global dataset covering $2021$-$2023$. The combination (logical OR) of predictions from these models resulted in a unified dataset. To ensure the dataset's quality and reliability, we implemented a rigorous quality control process. This included applying stringent filtering techniques to remove artifacts and isolated clusters and systematically eliminating any erroneous or untrustworthy data points. The resulting dataset, refined through these procedures and careful manual inspection, represents the latest and most accurate collection of data. Our comprehensive methodology ensures the delivery of a dataset distinguished by its precision and reliability, suitable for a broad spectrum of applications and analyses.

\section{Dataset Overview}
This section presents a detailed analysis of the proposed dataset, including comprehensive statistical information. It aims to give users a clear insight into the dataset's characteristics and strengths. Additionally, our Github repository contains an easy-to-use guide for working with this dataset regardless of the machine learning framework.


\subsection{Size, Region, and Time Span}\label{AA}
\begin{figure}[tb]
\centerline{\includegraphics[width=0.5\textwidth]{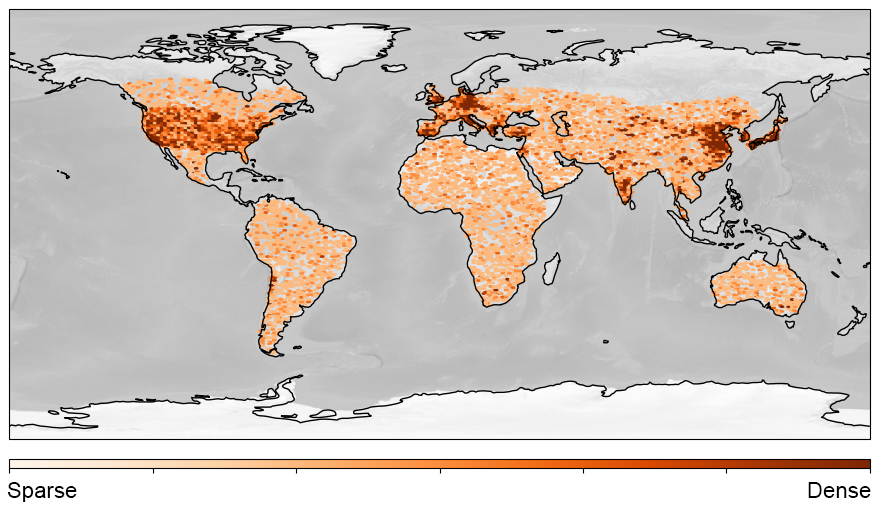}}
\caption{Global Distribution of Data Points. Sparse is represented in \textcolor{Apricot}{\textbf{light orange}}, while dense is represented in \textcolor{Brown}{\textbf{dark orange}}.}
\label{fig-data-distribution}
\end{figure}

The proposed dataset is comprised of a total of $13,703$ data samples, spanning various global regions and capturing information from the years $2021$ to $2023$. Fig. \ref{fig-data-distribution} displays the location and density of all data points on a global map. Furthermore, Table \ref{tab:data-comp} provides a detailed view of the dataset's composition and distribution

\begin{table}[tb]
\caption{Dataset Composition}
\begin{center}
\begin{tabular}{c|c|c|c}
\hline
\textbf{Year \& Region} & \textbf{Positive Sample}& \textbf{Negative Sample}& \textbf{Sum} \\
\hline
2023 US$^{\mathrm{*}}$&706&969&1675 \\
\hline
2022 US$^{\mathrm{*}}$&723&968&1691 \\
\hline
2023 Global&1774&1689&3463 \\
\hline
2022 Global&1803&1655&3458 \\
\hline
2021 Global&1787&1629&3416 \\
\hline\hline
\textbf{Total}&\textbf{6793}&\textbf{6910}&\textbf{13703} \\
\hline
\multicolumn{4}{l}{$^{\mathrm{*}}$Samples from the US region have been manually annotated.}
\end{tabular}
\label{tab:data-comp}
\end{center}
\end{table}

\subsection{Resolution and Dimension}

This dataset maintains a spatial resolution of $10$ meters per pixel, consistent with the resolution of Sentinel-2 satellite imagery. Each data sample within the dataset is standardized to dimensions of $256 \times 256$ pixels, and encompasses a rich spectral profile with $13$ bands \cite{sentinel}. Table \ref{tab:bands-info} summarizes the spectral information and Fig. \ref{fig-data-sample} presents two samples visualizing the $13$ bands along with the ground truth mask.

\begin{table}[tb]
    \caption{Dataset Bands Information}
    \centering
    \begin{tabular}{c|c|c}
        \hline
        \textbf{Name} & \textbf{Wavelength} & \textbf{Description} \\
        \hline
        B1 & 443.9nm (S2A) / 442.3nm (S2B) & Aerosols \\
        \hline
        B2 & 496.6nm (S2A) / 492.1nm (S2B) & Blue \\
        \hline
        B3 & 560nm (S2A) / 559nm (S2B) & Green \\
        \hline
        B4 & 664.5nm (S2A) / 665nm (S2B) & Red \\
        \hline
        B5 & 703.9nm (S2A) / 703.8nm (S2B) & Red Edge 1 \\
        \hline
        B6 & 740.2nm (S2A) / 739.1nm (S2B) & Red Edge 2 \\
        \hline
        B7 & 782.5nm (S2A) / 779.7nm (S2B) & Red Edge 3 \\
        \hline
        B8 & 835.1nm (S2A) / 833nm (S2B) & NIR \\
        \hline
        B8A & 864.8nm (S2A) / 864nm (S2B) & Red Edge 4 \\
        \hline
        B9 & 945nm (S2A) / 943.2nm (S2B) & Water vapor \\
        \hline
        B10 & 1373.5nm (S2A) / 1376.9nm (S2B) & Cirrus \\
        \hline
        B11 & 1613.7nm (S2A) / 1610.4nm (S2B) & SWIR 1 \\
        \hline
        B12 & 2202.4nm (S2A) / 2185.7nm (S2B) & SWIR 2 \\
        \hline
    \end{tabular}
    \label{tab:bands-info}
\end{table}

\subsection{Compare to Existing Datasets}\label{SCM}
Numerous studies are dedicated to examining datasets featuring high-resolution images of solar panels. SolarDK, a publicly available benchmark dataset, focuses on the geographical area of Denmark \cite{solardk}. This dataset encompasses human-annotated information for both classification and segmentation tasks. Furthermore, the UC Berkeley team developed another high-resolution dataset utilizing Google Maps to collect satellite images from various states in the U.S. \cite{hyperionsolarnet}. Although these openly accessible datasets are valuable resources, they lack diversity in covering non-urban regions and are restricted to specific geographic locations and countries. To address these shortcomings, our dataset overcomes these limitations by employing mid-resolution data and encompassing rural areas across global regions on various continents.

In $2021$, L. Kruitwagen created a dataset that offers a global inventory of photovoltaic solar energy generating units, presenting a comprehensive overview of solar panel distribution worldwide \cite{globalinventory}. This dataset involved an extensive analysis, processing $550$ terabytes of imagery spanning approximately $2$ months in real-time. While the outcome is an expansive and insightful resource, it has certain limitations. It exclusively provides shape files for solar panel areas, lacking crucial input source data. Moreover, it captures information within an outdated timeframe, spanning from $2016$ to $2018$. In response to these drawbacks, our proposed dataset aims to address these issues comprehensively. It will include multispectral input satellite images, offering a richer and more detailed dataset. Additionally, our dataset will cover an up-to-date timeframe, ranging from $2021$ to $2023$, ensuring a more current and relevant depiction of the global solar energy landscape.

\section{Benchmark Models and Results}

\begin{figure*}[t]
\centerline{\includegraphics[width=\textwidth]{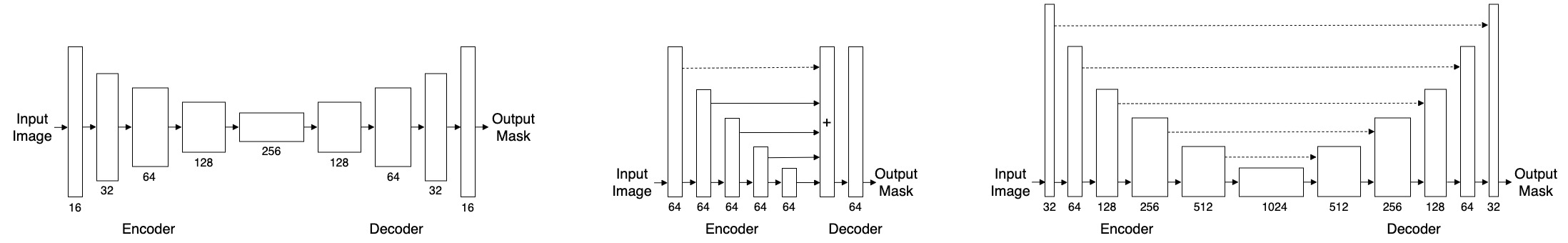}}
\caption{Architecture of FCN, Half-UNet, and U-Net}
\label{fig-fcn-unet-arch}
\end{figure*}

Running benchmark segmentation models on our dataset is essential to establish a performance baseline and enable a fair comparative evaluation. This not only enhances our understanding of the dataset but also provides a standardized reference for the broader research community, fostering transparency, consistency, and credibility in future studies. To analyze and benchmark the dataset, we employed two widely recognized semantic segmentation models and one simplified model, namely the Fully Convolutional Network (FCN), U-Net, and Half-UNet depicted in Figure \ref{fig-fcn-unet-arch}.


\begin{table}[t]
\caption{Benchmark Model Results}
\centering
\begin{tabular}{lccc}
\hline
\textbf{Model} & \textbf{Params} & \textbf{IoU (\%)} & \textbf{F-score (\%)} \\
\hline
FCN\cite{fcn} & 1,996,737 & 71.81 & 82.87 \\
Half-UNet\cite{halfunet} & 224,609 & 71.48 & 82.17 \\
U-Net\cite{unet} & 31,129,377 & 79.31 & 87.80 \\
\hline
\end{tabular}
\label{tab:seg_results}
\vspace{-2mm}
\end{table}

\begin{figure*}[t]
\centerline{\includegraphics[width=\textwidth]{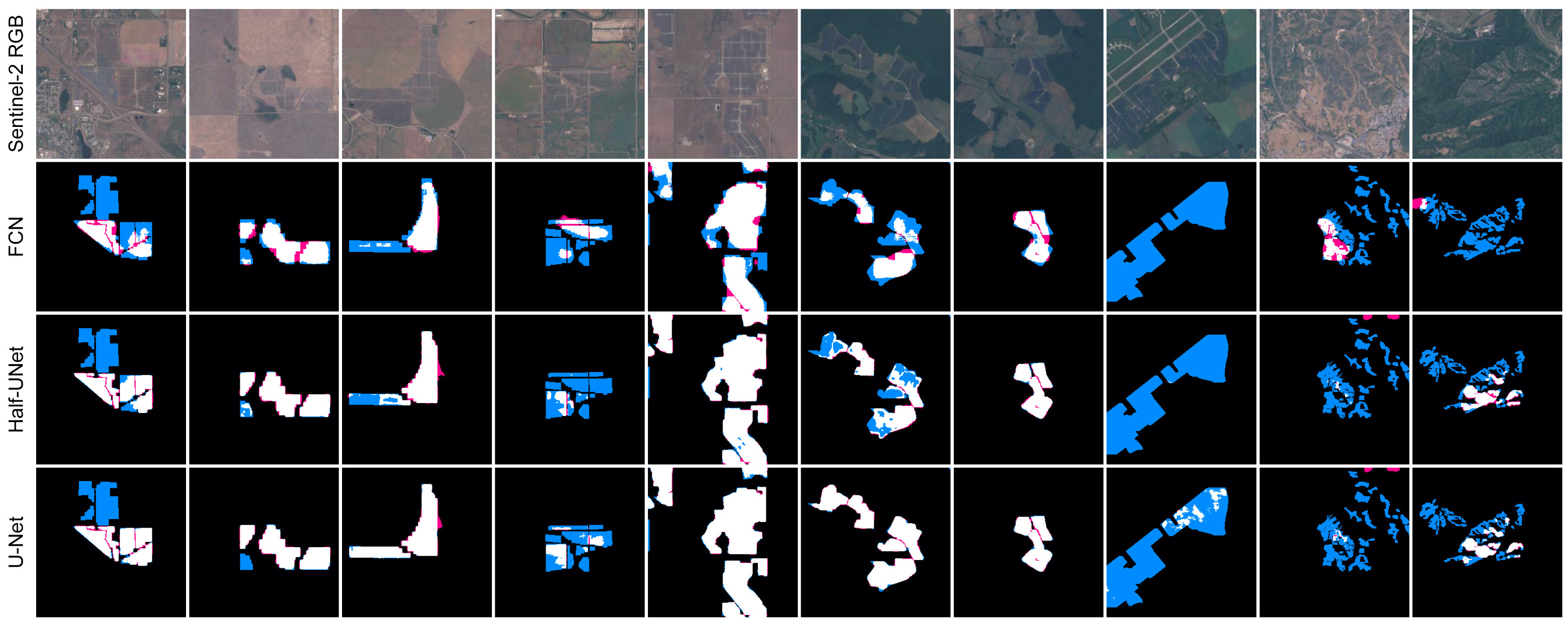}}
\caption{Qualitative Results of FCN, Half-UNet, and U-Net Models. The first row displays the test samples in true color. Rows $2$-$4$ depict the prediction masks for solar farm areas. In this representation, True Positive is shown as \textbf{white}, True Negative as \textbf{black}, False Positive as \textcolor[rgb]{1,0,0.55}{\textbf{red}}, and False Negative as \textcolor[rgb]{0,0.55,1}{\textbf{blue}}.}
\label{fig-fcn-unet-result}
\vspace{-2mm}
\end{figure*}

The FCN, introduced by Long et al. \cite{fcn}, is renowned for its ability to perform end-to-end pixel-wise prediction through the extensive utilization of convolutional layers, allowing it to maintain spatial information throughout the network. This makes FCN particularly effective for semantic segmentation tasks.On the other hand, the U-Net architecture, proposed by Ronneberger et al. \cite{unet}, features a U-shaped structure that incorporates both encoder and decoder components. The encoder captures contextual information by progressively downsampling the input, while the decoder facilitates precise localization by upsampling and merging features from different resolutions. The Half-UNet \cite{halfunet}, a simplified model, offers a more lightweight alternative. While retaining the essence of U-Net, Half-UNet reduces computational complexity by utilizing only half of the U-shaped architecture. This streamlining allows for faster processing and lower resource requirements, making it an attractive choice for scenarios with limited computational resources. 

For the training of three models, a batch size of $16$ was selected to balance computational efficiency and model convergence, and training was conducted over $100$ epochs. To mitigate the risk of overfitting, we implemented early stopping with a patience value set to $10$ based on validation loss. These hyperparameter choices were made to strike a balance between model complexity and training efficiency. The summary of the performance metrics and model size are presented in TABLE \ref{tab:seg_results}. Additionally, Fig. \ref{fig-fcn-unet-result} illustrates the qualitative results of the three models.


\vspace{-2mm}
\section{Conclusion}
This study presented a comprehensive global dataset of multispectral satellite images of solar panel farms. Our dataset enhances existing resources in three ways: by incorporating mid-resolution data from rural areas globally, using multispectral satellite images for richer detail, and updating the timeframe of data to $2021$-$2023$ for a current view of the global solar energy landscape. The evaluation of this dataset using benchmark models has established its robustness and set a baseline for future improvements. It has proven effective in accurately identifying solar panel areas on a global scale. This dataset is a significant asset for the solar energy research community, potentially driving advancements in renewable energy development.

\vspace{-4mm}
\medskip
\bibliographystyle{IEEEtranN}
\bibliography{references}
\end{document}